\documentclass[letterpaper, 10 pt, conference]{ieeeconf}  

\IEEEoverridecommandlockouts                              

\usepackage{graphics} 
\usepackage{epsfig} 
\usepackage{mathptmx} 
\usepackage{times} 
\usepackage{amsmath} 
\usepackage{amssymb}  
\usepackage{tabularx,ragged2e,booktabs,caption}
\usepackage{url}            
\usepackage{booktabs}       
\usepackage{amsfonts}       
\usepackage{nicefrac}       
\usepackage{microtype}      
\usepackage[linesnumbered,ruled,vlined]{algorithm2e}
\usepackage{sidecap}
\usepackage{wrapfig}
\usepackage{lipsum}
\usepackage{subfigure}
\usepackage{color}
\usepackage[sort,compress]{cite}

\usepackage[linesnumbered,ruled,vlined]{algorithm2e}
\newcommand{\be}{\begin{equation}}
\newcommand{\ee}{\end{equation}}

\DeclareMathOperator*{\argmin}{\arg\!\min}
\DeclareMathOperator*{\argmax}{\arg\!\max}

\title{\bf Context-Aware Pedestrian Motion Prediction In Urban Intersections}

\author{Golnaz Habibi\textsuperscript{*}, Nikita Jaipuria\textsuperscript{*}, Jonathan P. How\\
Massachusetts Institute of Technology\\
\{golnaz, nikitaj, jhow\}@mit.edu\\
\thanks{*These authors contributed equally and their names are in alphabetical order.}
}

\begin{document}

\maketitle
\thispagestyle{empty}
\pagestyle{empty}

\begin{abstract}
This paper presents a novel context-based approach for pedestrian motion prediction in crowded, urban intersections, with the additional flexibility of prediction in similar, but new, environments. Previously, \cite{chen2016augmented} combined Markovian-based and clustering-based approaches to learn motion primitives in a grid-based world and subsequently predict pedestrian trajectories by modeling the transition between learned primitives as a Gaussian Process (GP). This work extends that prior approach by incorporating semantic features from the environment (relative distance to curbside and status of pedestrian traffic lights) in the GP formulation for more accurate predictions of pedestrian trajectories over the same timescale. We evaluate the new approach on real-world data collected using one of the vehicles in the MIT Mobility On Demand fleet \cite{miller2017predictive}. The results show 12.5\% improvement in prediction accuracy and a 2.65$\times$ reduction in Area Under the Curve (AUC), which is used as a metric to quantify the span of predicted set of trajectories, such that a lower AUC corresponds to a higher level of confidence in the future direction of pedestrian motion. 
\end{abstract}

\section{INTRODUCTION}
Recent advances in sensor technologies and computing power have led to a surge in research on autonomous driving to improve road safety (\cite{fagnant2015preparing,Bagloee2016}), reduce traffic congestion and improve vehicle utilization. For safe and efficient autonomous driving in complex urban environments, a self-driving vehicle must be able to interact with other moving objects, including pedestrians, cyclists, and, of course, cars. Pedestrian trajectory prediction is challenging as compared to that of other cars and cyclists because of the absence of a regular flow, such as driving within lanes and staying within road boundaries, that results from a fairly uniform set of predefined ``rules of the road'' for cars (and to some extent cyclists). The complexity is increased further when the urban environment includes pedestrian traffic lights or tightly packed sidewalks with numerous pedestrian interactions. There is a need for an efficient pedestrian motion prediction algorithm that can address these challenges.

\begin{figure}[t]
	\begin{center}
		\includegraphics[width=.5\textwidth]{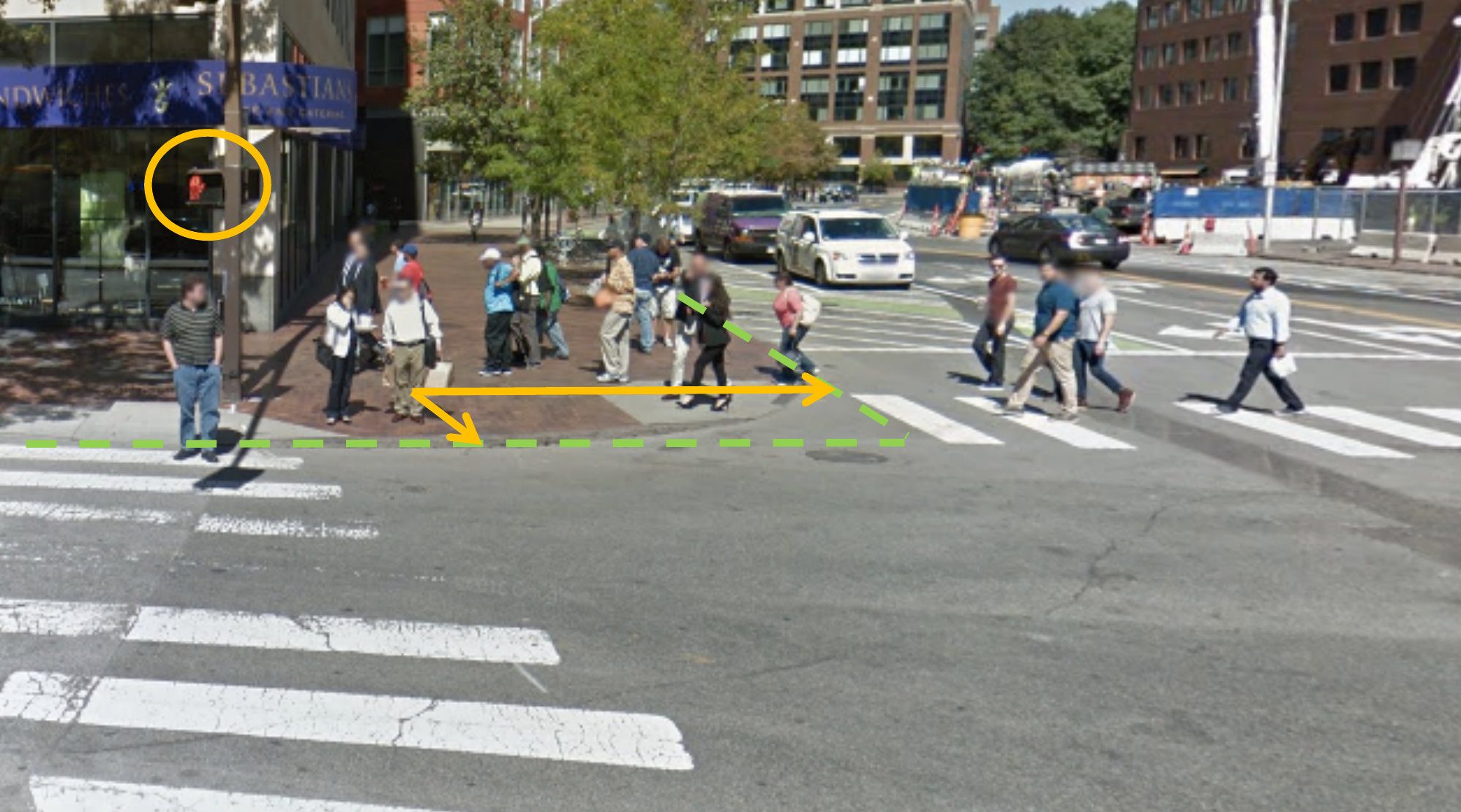} 
	\end{center}
	\caption{\small Example intersection scenario. Dotted green line denotes a rectangular approximation to the curbside in view. Orange arrows denote relative distance of a pedestrian from the two curbsides, which can indicate pedestrian intention. Pedestrian traffic light status is highlighted in orange, which also influences pedestrian movement.
	\label{fig:intersection}}
\end{figure}
Most of the previous work on mobile agent trajectory prediction is either prototype-trajectory based or Markovian maneuver intention estimation-based~\cite{lefevre2014survey}. \cite{chen2016augmented} use a combination of the two, to inherit the benefits of both, in developing a dictionary learning algorithm, called augmented semi nonnegative sparse coding (ASNSC). Learning motion primitives instead of complete prototype trajectories addresses the issue of partial observability of trajectories caused by occlusions or a limited field of view of on-board perception sensors. ASNSC outputs a set of feasible trajectories as its prediction that are learned based on solely the spatial features of the training dataset (absolute $x$ and $y$ position and orientation of pedestrians), ignoring the environment context that may influence a pedestrian's intent.

The accuracy of these predictions could be improved by adding semantic features from the environment in the learning process. Incorporating context can also provide flexibility of application of the learned model to prediction in new, but similar environments, unexplored earlier, which is in general difficult to achieve with clustering-based approaches~\cite{lefevre2014survey}. Fig.~\ref{fig:intersection} shows an intersection scenario in which, when faced by a choice between two crosswalks, pedestrian traffic light status for each of those crosswalks influences pedestrian choice. Similarly, a comparison of the relative distance to each curbside could be indicative of future direction of motion. 

This paper extends~\cite{chen2016augmented} by incorporating semantic features from the environment. To meet this objective, a dictionary of motion primitives is learned using ASNSC, as in~\cite{chen2016augmented}. However, the GP based modeling of transition between learned motion primitives is done with respect to the environmental context instead of absolute x and y position. As illustrated in Fig.~\ref{fig:dictionary_steven}, the environmental context, such as pedestrian traffic light status, influences the probability of transition between two motion primitives. This aspect, however, was not captured in ASNSC. The two main context features used in this work are pedestrian traffic light status and relative distance to curbside. A squared exponential (SE) kernel function with automatic relevance determination (ARD)~\cite{rasmussen2006gaussian} is used to determine the relevance of each of the individual features. Previously, we presented some initial results in the 2017 NIPS Machine Learning for Intelligent Transportation Systems workshop. This paper builds on that previous work by introducing three different context feature sets for learning the transition between motion primitives. An extensive evaluation of the prediction performance of our context-aware approach with all three feature sets is also provided. A quantitative comparison of our approach with ASNSC shows a $12.5\%$ increase in the prediction accuracy.

The main contributions of this work are:
\begin{itemize}
\item Context-based ASNSC (CASNSC) framework for embedding context such as traffic lights, relative distance to curbside etc. in~\cite{chen2016augmented};
\item Comparison of three different variations of CASNSC with ASNSC to show improvement in prediction accuracy;
\item Utilization of context to transform pedestrian position in the global $x-y$ coordinate frame into a rotated $x'-y'$ coordinate frame, in which the two coordinates are independent of each other (see Fig.~\ref{fig:curbside}). This aids in building more accurate GPs for modeling the transition between learned motion primitives;
\item Selection of context features that are invariant to the training intersection geometry (for orthogonal intersections). This lays the foundations for trajectory prediction using CASNSC in intersections other than the one it has been trained on.
\end{itemize}

\section{RELATED WORK}
Several papers have been written on short-term prediction of human motion~\cite{kooij2014context,bissacco2009hybrid,gonzalez2004context,goldhammer2013early}, but understanding goals or intent is needed to plan for longer timescales~\cite{karasev2016intent,alahi2016social}. For instance,~\cite{jacobs2017real} demonstrates the ability to accurately predict the final destination of pedestrians using a probabilistic pedestrian modeling approach. Our aim, however, is to not just predict the final destination, but also the path taken by the pedestrian to reach its goal.  Previous work has focused on two main approaches for trajectory prediction~\cite{lefevre2014survey}: prototype trajectories-based and  maneuver intention estimation-based. In general, prototype trajectories-based approaches are more robust to measurement noise when compared to maneuver intention estimation-based approaches, which are mostly Markovian~\cite{makris2002spatial,vasquez2009incremental,schulz2015controlled} and therefore, rely on the current state only for prediction. However, the prototype trajectories-based approaches can be computationally quite expensive~\cite{rasmussen2002infinite,ferguson2015real} and hence slow in detecting changes in pedestrian intent. They are also susceptible to issues like partial trajectories in the training dataset being grouped together into a cluster and learned as a trajectory prototype. \cite{chen2016augmented} combine these two approaches to inherit the benefits of both in developing ASNSC. They achieved significant improvement over state-of-art clustering based approach using Dirichlet Process mixture of Gaussian Process (DPGP). However, an important limitation of~\cite{chen2016augmented} is that available environmental context is not utilized for trajectory prediction.

Most of the previous work on context-based pedestrian trajectory prediction is limited to a classification problem~\cite{schulz2015pedestrian, gonzalez2004context, schneemann2016context}. In addition, some are also based on the limiting assumption of only one context feature being active at a time, which works for short-term, immediate prediction only~\cite{gonzalez2004context}. More recently,~\cite{schulz2015controlled} use a combination of an Interacting Multiple Model (IMM) filter for tracking and Latent-dynamic Conditional Random Field (LDCRF) model for intention prediction. Their approach implicitly utilizes situational awareness by embedding human head pose into the LDCRF model and the prediction horizon is limited to 1 second. Our model, in contrast, predicts on explicit inclusion of context, for a long-term prediction horizon of 5 seconds. Further,~\cite{karasev2016intent} used jump-Markov process for long term prediction of pedestrian motion by incorporating traffic light and crosswalks as semantic features. The output of their prediction model is an \emph{occupancy map} of feasible trajectory predictions. Our goal instead is to make prediction confident and output the \emph{most likely trajectory} with increased accuracy by incorporating context in the ASNSC based prediction model~\cite{chen2016augmented}.
\section{Background}

In this section, we first briefly review the ASNSC algorithm for learning motion primitives followed by a review of the GP based framework for trajectory prediction using the learned dictionary, as in~\cite{chen2016augmented}.

\subsection{Augmented Semi-Nonnegative Sparse Coding}

Given a training dataset of $n$ samples, $\mathbf{Z} = [\mathbf{x_1}, \ldots, \mathbf{x}_n]$, where $\mathbf{x_i}$ is a column vector of length $p$, the objective is to learn a set of $K$ dictionary atoms, $\mathbf{D} = [\mathbf{d_1}, \ldots, \mathbf{d}_K]$, and the corresponding nonnegative sparse coefficients, $\mathbf{S} = [\mathbf{s_1}, \ldots, \mathbf{s}_n]$. Mathematically, this can be formulated as a constrained optimization problem of the form~\cite{chen2016augmented}:
\begin{align}
\argmin_{\mathbf{D}, \mathbf{S}} & \;  \left| \left| \mathbf{Z} - \mathbf{DS} \right| \right|_F^2 + \lambda \sum \limits_{i=1}^n \left| \left| \mathbf{s_i} \right |\right|_1  \label{eq::dictionary} \\
\text{s.t. } & \mathbf{d_k} \in \mathbf{Q}, \ s_{ki} \geq 0 \ \forall \ \text{ k,i} \nonumber
\end{align} 
where $\lambda$ is a regularization parameter and $\mathbf{Q}$ is the feasible set in which $\mathbf{d_k}$ resides. In particular, this framework can be applied to learning a dictionary of motion primitives for pedestrian trajectories.

\begin{figure}[h]
\centering
  \subfigure[]{\label{fig:grid}
  \includegraphics[width=0.2\linewidth]{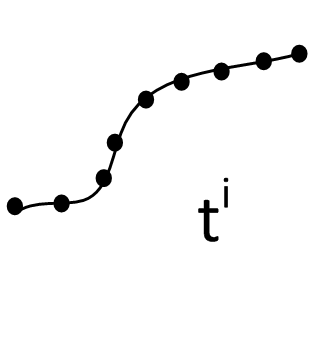}
  \quad
  \includegraphics[width=0.18\linewidth]{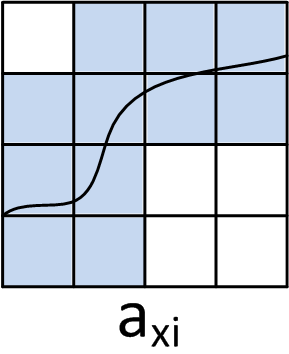}
  \quad
  \includegraphics[width=0.4\linewidth]{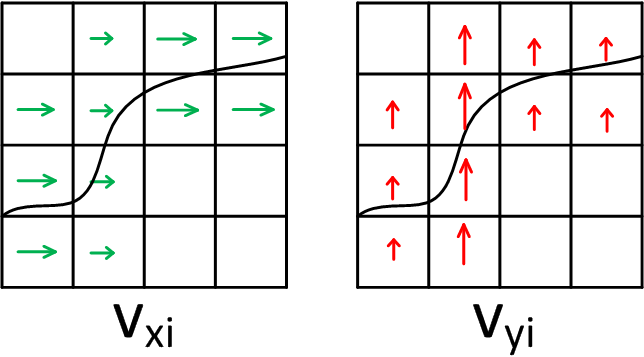}}
  \subfigure[]{\label{fig:dictionary_steven}
  \includegraphics[width=0.45\linewidth]{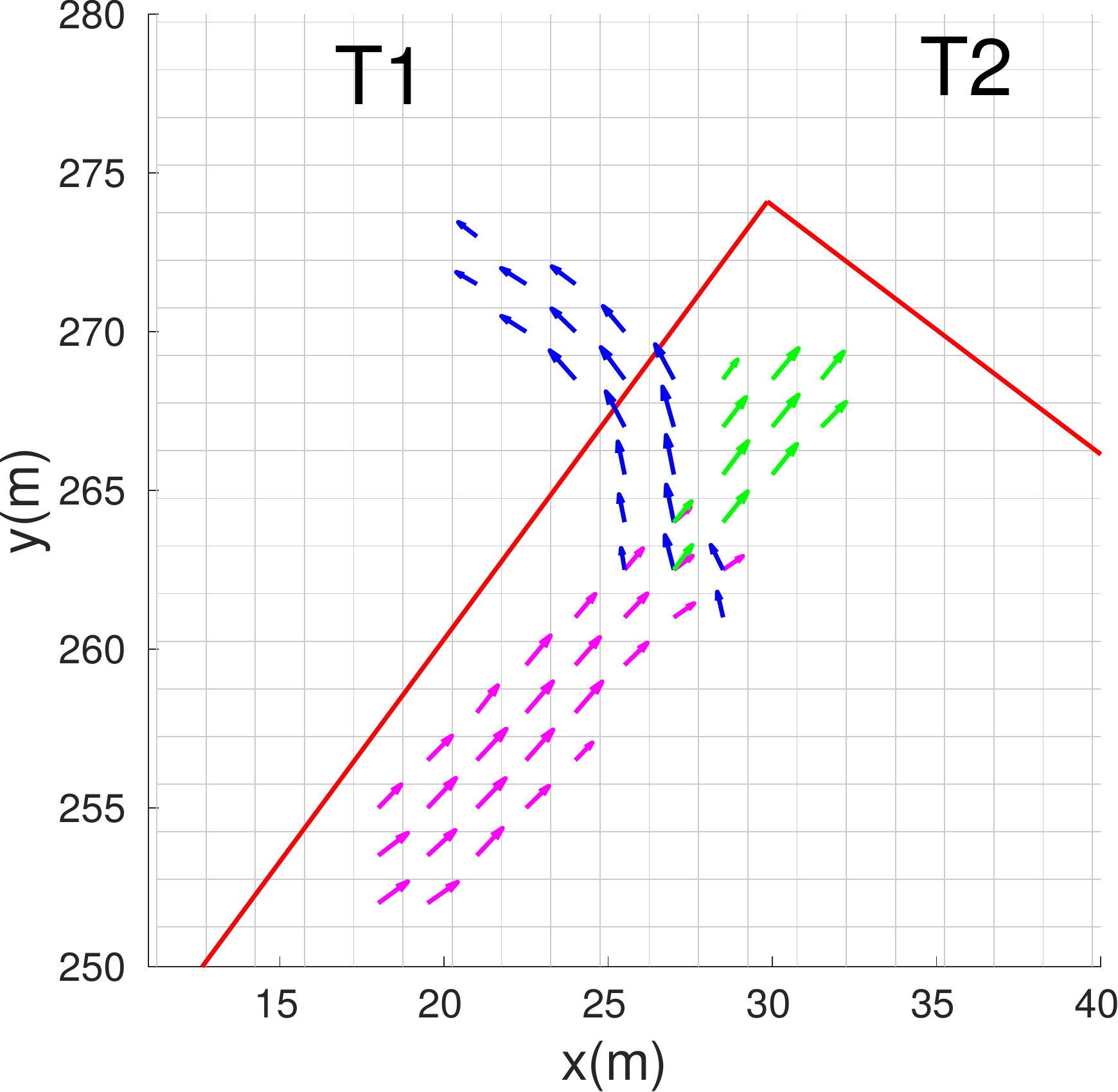}
  \quad
  \includegraphics[width=0.45\linewidth]{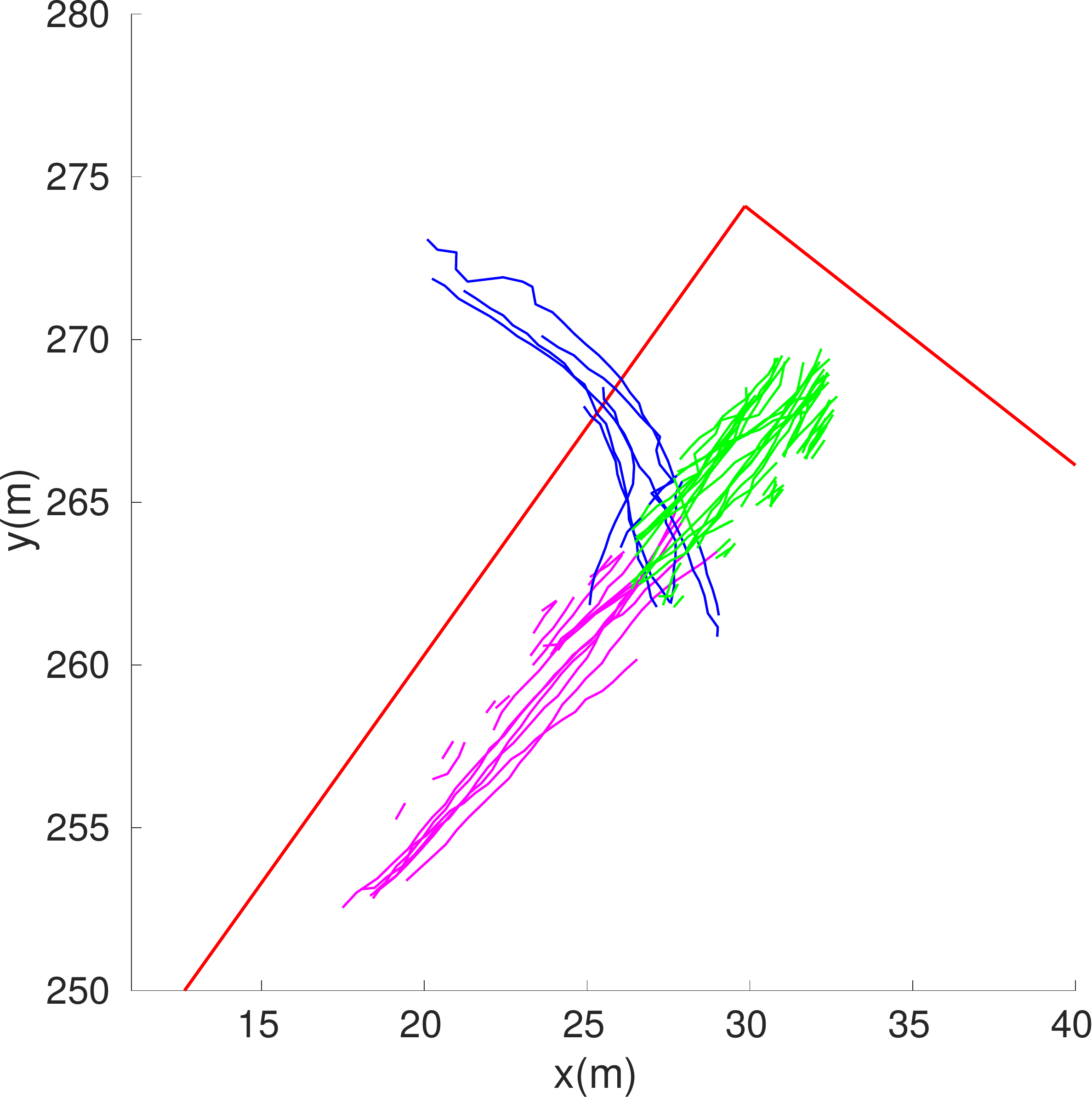}}
  \quad
\caption{\small (a) From~\cite{chen2016augmented}, an illustration of the vector representation of a trajectory $t_i$ as $\mathbf{x_i}^T = [\mathbf{v_{x_i}}^T, \mathbf{v_{y_i}}^T, \mathbf{a_i}^T]$, where $\mathbf{v_{x_i}},\mathbf{v_{y_i}} \in \mathbb{R}^{MN}$ represent the normalized x, y velocities in a grid-based world and $\mathbf{a_i}^T$ are the activeness variables. (b) An example of dictionary atoms/motion primitives learned using the ASNSC framework (left) and clustering of training trajectories on the basis of the learned dictionary atoms for computing the transition matrix $\mathbf{T}$ (right). Each dictionary atom is shown in a different color. T1 and T2 denote two different traffic lights, the status of which influences transition between dictionary atoms. For e.g., the transition between dictionary atoms shown in magenta and blue has a higher probability than that between dictionary atoms shown in magenta and green for T1 = 1 (crosswalk clear for pedestrians to cross), T2 = 0}
  \label{fig:grid_dictionary}
\end{figure}

For this particular application, the input consists of $n$ pedestrian trajectories, where each trajectory $t_i$ is a sequence of two-dimensional position measurements taken at a fixed time interval $\Delta t$. The output is a dictionary of motion primitives $\mathbf{D}$; an example is shown in Fig.~\ref{fig:dictionary_steven}. As described in~\cite{chen2016augmented}, a discretized world, consisting of $M\times N$ blocks of width $w$, is used to develop a vector representation $\mathbf{x_i}$ of each training trajectory $t_i$ as shown in Fig.~\ref{fig:grid}. Since we are interested in just the shape of predicted trajectory, the input velocities are normalized.

\subsection{Trajectory prediction using the learned dictionary}
As shown in Fig.~\ref{fig:dictionary_steven}, $\mathbf{D}$ is used to segment the original training trajectories $\mathbf{x_i}$ into clusters, where each cluster is best explained by one of the learned dictionary atoms. A transition matrix, $\mathbf{T} \in \mathbb{Z}^{K\times K}$ is thus created, where $T(i,j)$ denotes the number of trajectories exhibiting a transition from the $i$-th dictionary atom to the $j$-th dictionary atom. A transition will, therefore, be mathematically represented as a concatenation of two dictionary atoms $\{\mathbf{d_i}, \mathbf{d_j}|T(i,j)>0\}$. Each transition is modeled as a two-dimensional GP flow field~\cite{joseph2011bayesian,aoude2013probabilistically}. In particular, two independent GPs, $(GP_x,GP_y)$, called GP motion patterns, are used to learn a mapping from the chosen features $\mathbf{X} \in \mathbb{R}^N$, where $N$ denotes the number of features, to the x-y velocities.
\begin{eqnarray}
GP_x: \mathbf{X} \to v_x, \quad GP_y: \mathbf{X} \to v_y
\label{eq:gp1}
\end{eqnarray}
ASNSC uses $\mathbf{X} = \mathbf{X_p} = (x,y)^T \in \mathbb{R}^2$ as the feature vector. The learned GP motion patterns, $(GP_x,GP_y)$, are used for generating a predicted path using (\ref{eq:gp1}) as well as for computing the likelihood of an observed trajectory, $\mathbf{t}' = \{(\mathbf{X_1}',\mathbf{v_1}'),\ldots,(\mathbf{X_l}',\mathbf{v_l}')\}$ using 
\begin{align}
\label{eq:old4}
	P(\mathbf{t}'|GP_x,GP_y) = &\prod_{\mathbf{X}'\in \mathbf{t}'} \mathcal{N}(v_x;\mu_{GP_x}(\mathbf{X}'),\sigma^2_{GP_x}(\mathbf{X}'))\\
									 & \mathcal{N}(v_y;\mu_{GP_y}(\mathbf{X}'),\sigma^2_{GP_y}(\mathbf{X}')) \nonumber
\end{align}

Trajectory prediction has two main steps. 1) Unitary GP motion patterns, $(GP^{\ uni}_x,GP^{\ uni}_y)$, are learned from training trajectories corresponding to $\mathbf{T}(i,j) \ \forall \ i=j$. The unitary GP motion pattern that most likely generated the observed trajectory $\mathbf{t}'$ is determined using (\ref{eq:old4}), which is equivalent to selecting the most likely initial dictionary atom $\mathbf{d}_{\hat{k}}$ (Algorithm~\ref{alg:Algorithm1}, line 11). 2) The set of possible future dictionary atoms can be found as $\mathcal{D} = \{j|\mathbf{T}_{\hat{k}j}>0\}$ (Algorithm~\ref{alg:Algorithm1}, line 12). Transitional GP motion patterns, $(GP^{\ tran}_{x_{\hat{k}j}},GP^{\ tran}_{y_{\hat{k}j}}) \ \forall j \in \mathcal{D}$ are then used for generating a set of predicted trajectories $\mathbf{s}_j$. 

\cite{chen2016augmented} provides an improved prediction of pedestrian trajectories as compared to that using Gibbs sampling for DPGP~\cite{ferguson2015real}. However, the environmental context is not taken into account when performing prediction. This paper presents CASNSC as an extension of~\cite{chen2016augmented} by embedding context into their prediction model. The following section introduces the context features used and our approach for incorporating them into the trajectory prediction model of~\cite{chen2016augmented}.

\section{ALGORITHM}
As discussed earlier, this work extends ASNSC by incorporating semantic features from the environment in the transition learning phase (Algorithm~\ref{alg:Algorithm1}, lines 5-13) and is motivated by situations in which context influences transition between learned dictionary atoms (see Fig.~\ref{fig:dictionary_steven}). The proposed approach uses two sets of features: 1) \emph{dictionary features}, $\mathbf{X_d}$, which are used for learning the dictionary $\mathbf{D}$ (Algorithm~\ref{alg:Algorithm1}, lines 1-4); and 2) \emph{transition features}, $\mathbf{X_t}$, which are used for learning the transition between dictionary atoms using GP models (Algorithm~\ref{alg:Algorithm1}, lines 5-13). ASNSC uses the same set of two-dimensional position feature, $(x,y)^T$, as both $\mathbf{X_d}$ and $\mathbf{X_t}$. We propose three different feature sets as $\mathbf{X_t}$ for learning the GP models, such that each feature set is increasingly less dependent on pedestrian position and more dependent on context instead.

\subsection{Context features}
\subsubsection{Pedestrian traffic light}
A pedestrian's decision to go left or right is influenced by the status of two pedestrian traffic lights (T1, T2) in a four-way intersection scenario. A single-dimensional feature vector, $(tr)$, is sufficient to capture the environment context with respect to both the traffic lights as the change in status of (T1, T2) captures redundant information.

\subsubsection{Curbside orientation}
Pedestrian motion in sidewalks is constrained by the orientation of the curbsides. An arbitrarily chosen x-y coordinate frame, therefore, results in a dependence of the coordinates on each other. This violates the assumption of independence of x, y coordinates in the SE kernel function used by the GP models. As shown in Fig.~\ref{fig:curbside}, rotating the $x-y$ frame into the $x'-y'$ frame, which has the same orientation as that of the curbsides of interest can reduce the dependence of the coordinates on each other. Such a transformation can improve GP modelling, and consequently, trajectory prediction accuracy. The described transformation is equivalent to embedding the curbside/sidewalk orientation as a context feature in the prediction model.

\subsubsection{Relative distance to curbside}
\begin{figure}[t]
\centering
  \includegraphics[width=1\linewidth]{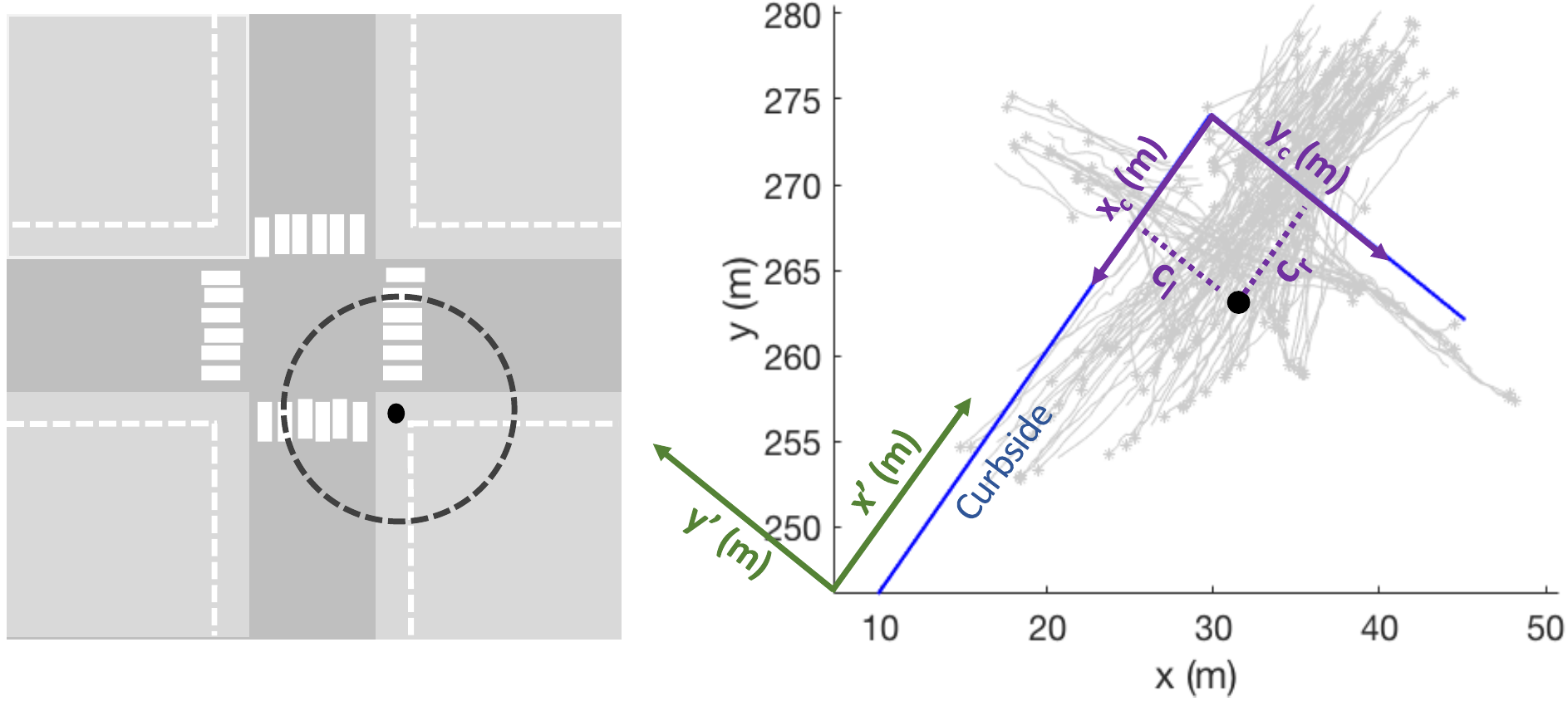}
  \caption{\small A typical four-way intersection (left) is used to explain the \emph{curbside orientation} and \emph{relative distance to curbside} context features. The zoomed portion (right) shows a pedestrian location as a black dot. $(c_l,c_r)^T$ denotes the vector of distance to the two curbsides of interest and is used as the \emph{relative distance to curbside} context feature. The signs of vector elements $c_l$ and $c_r$ are determined using the curbside coordinate frame $x_{c}-y_{c}$. Pedestrian position in the rotated coordinate frame $x'-y'$, which has the same orientation as that of the curbside in the global coordinate frame $x-y$, is used as the \emph{curbside orientation} context feature.}
    \label{fig:curbside}
\end{figure}

In addition to the curbside orientation, the relative distance of a pedestrian (treated as a point mass) to the curbside also provides useful contextual information. This distance can be computed using either a prior map of the environment or by online curb identification and localization. As, shown in Fig.~\ref{fig:curbside}, a two-dimensional vector, ${(c_l,c_r)}^T$ is used as the relative distance to curbside feature, which is equivalent to transforming the arbitrarily chosen global $x-y$ coordinate frame into the $x_c-y_c$ coordinate frame, that is exactly aligned with the curbsides of interest.

\subsection{Feature sets}
\subsubsection{Position and pedestrian traffic light}
The first feature set is a combination of the two-dimensional pedestrian position and the \emph{pedestrian traffic light} context feature, i.e., $\mathbf{X_t} = (x,y,tr)^T$. Application of the CASNSC framework with this particular feature set will be referred to as CASNSC-1.

\subsubsection{Curbside orientation and pedestrian traffic light}
As described earlier, an inherent limitation of the first feature set is the fact that x, y are not independent of each other in intersections as they are constrained by the geometry of the sidewalk/curbsides(see Fig.~\ref{fig:curbside}). This violates the x-y independence assumption made in the GP transition models. To address this issue, \emph{curbside orientation} is combined with the \emph{pedestrian traffic light} context feature to create another feature set $\mathbf{X_t} = (x',y',tr)^T$. The specific application of CASNSC with this feature set will be referred to as CASNSC-2.

\subsubsection{Relative distance to curbside and pedestrian traffic light}
Another important piece of contextual information missing in the second feature set is the actual location of the intersection corner/curbsides, which can also be an important indicator of pedestrian intent (see Fig.~\ref{fig:intersection}). We incorporate this missing piece of information by combining the \emph{relative distance to curbside} with the \emph{pedestrian traffic light} context feature to create the third feature set $\mathbf{X_t} = (c_l,c_r,tr)^T$. The CASNSC framework with this feature set will be referred to as CASNSC-3.

\begin{algorithm}[t]
    \tcc{Dictionary Learning Phase}
      	  $\mathbf{D} \leftarrow \mathbf{0}, \mathbf{S} \leftarrow \mathbf{0}$\;
          \While{not converged}
           {$\{\mathbf{D},\mathbf{S}\} = ASNSC(\mathbf{Z}, \mathbf{X_d}, \lambda)$}
          $\mathbf{T} \leftarrow \text{Transition\_Matrix} (\mathbf{D},\mathbf{Z},\mathbf{S})$\;
          \tcc{Transition Learning Phase}
          $GP^{\ uni} \leftarrow \emptyset , GP^{\ tran} \leftarrow \emptyset$\;
          \For{ $ \forall \ (i,j) ~\text{s.t.}~ \{ \mathbf{T}(i,j) > 0 \}$}{
          \If {$i == j$}
          {$GP^{\ uni}.insert((GP_x(\mathbf{X_t}),GP_y(\mathbf{X_t})))$ 
          }
          \Else 
          {
          	$GP^{\ tran}.insert((GP_x(\mathbf{X_t}),GP_y(\mathbf{X_t})))$
          }
          }
          $\hat{k} = \argmax_k \ P(\textbf{t}'|GP^{\ uni}_{x_{k}},GP^{\ uni}_{y_{k}})$ 

           \For{ $\forall \ j \ \in \mathcal{D} = \{ j|\mathbf{T}_{\hat{k}j}>0\}$ }{
          $\mathbf{s}_j \leftarrow \text{Predict}(\mathbf{t}',(GP_{x_{\hat{k}j}}^{\ tran},GP_{y_{\hat{k}j}}^{\ tran}))$}
          \caption{CASNSC - Context-based Augmented Semi Nonnegative Sparse Coding}
          \label{alg:Algorithm1}
\end{algorithm}

\subsection{Kernel function}
A SE kernel function with ARD is used as it allows for the combination of features with different characteristics and scales each feature in accordance with its relevance (\cite{rasmussen2006gaussian}).
\begin{eqnarray}
\small
k(\mathbf{X},\mathbf{X}') = \sigma_f^2 \exp(- \sum_{i=1}^{m}\frac{1}{2{l_{i}}^2}(x_i-x_i')^2)
\end{eqnarray}
where, $x_i \in \mathbf{X_t} ~\forall~ i = \{1,...,m\}$ is the i-th \emph{transition feature} and $l_{i}$ is the characteristic length of this feature. For instance, for predictions using CASNSC-1 where $\mathbf{X_t} = (x,y,tr)^T$, the hyper-parameters that need to be tuned would be given by the column vector $h = (l_x,l_y,l_{tr},\sigma_f)^T$. 
\section{RESULTS}
\begin{figure*}[h]
\centering
\subfigure[]{
\includegraphics[width=0.245\linewidth]{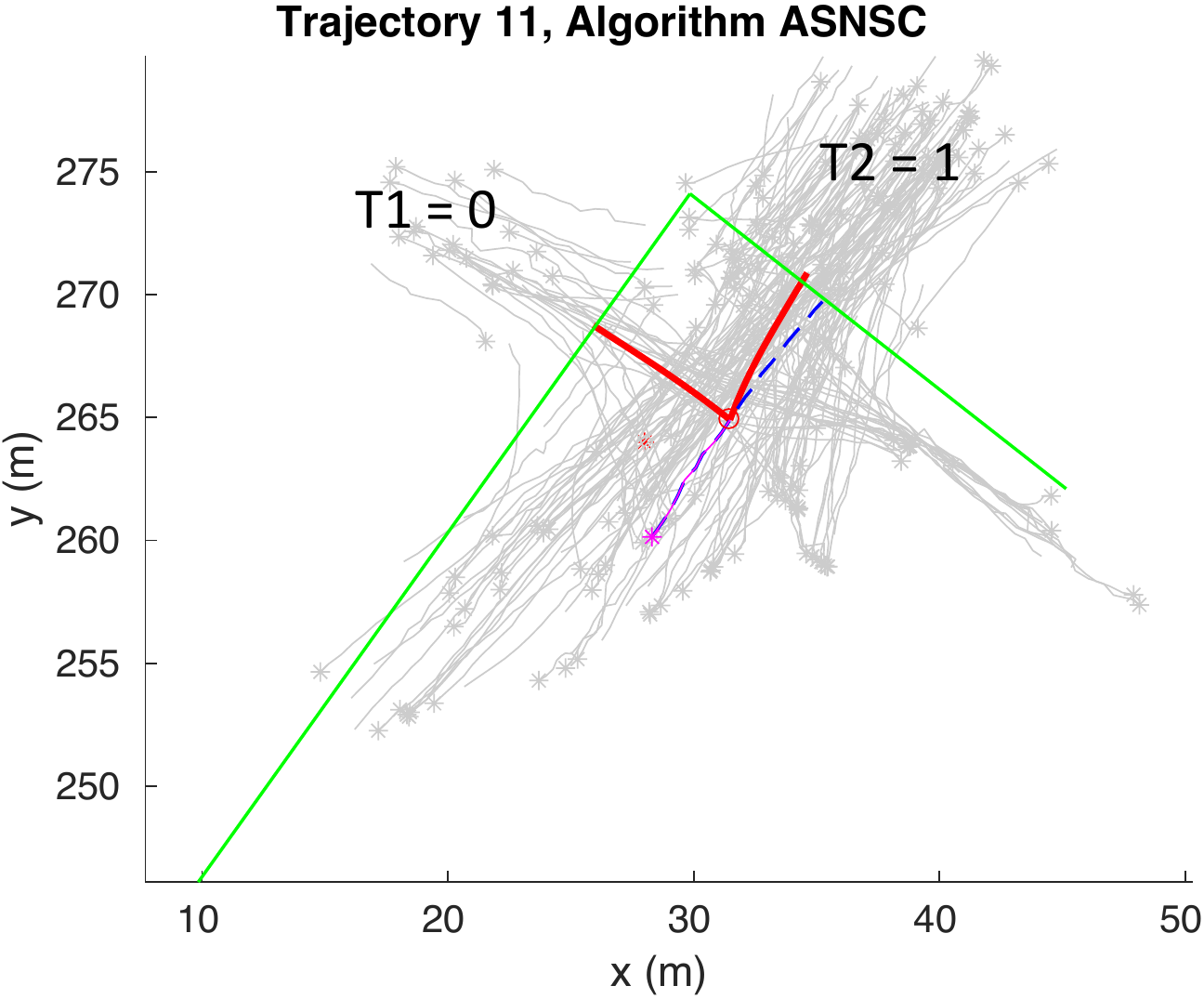}
\includegraphics[width=0.245\linewidth]{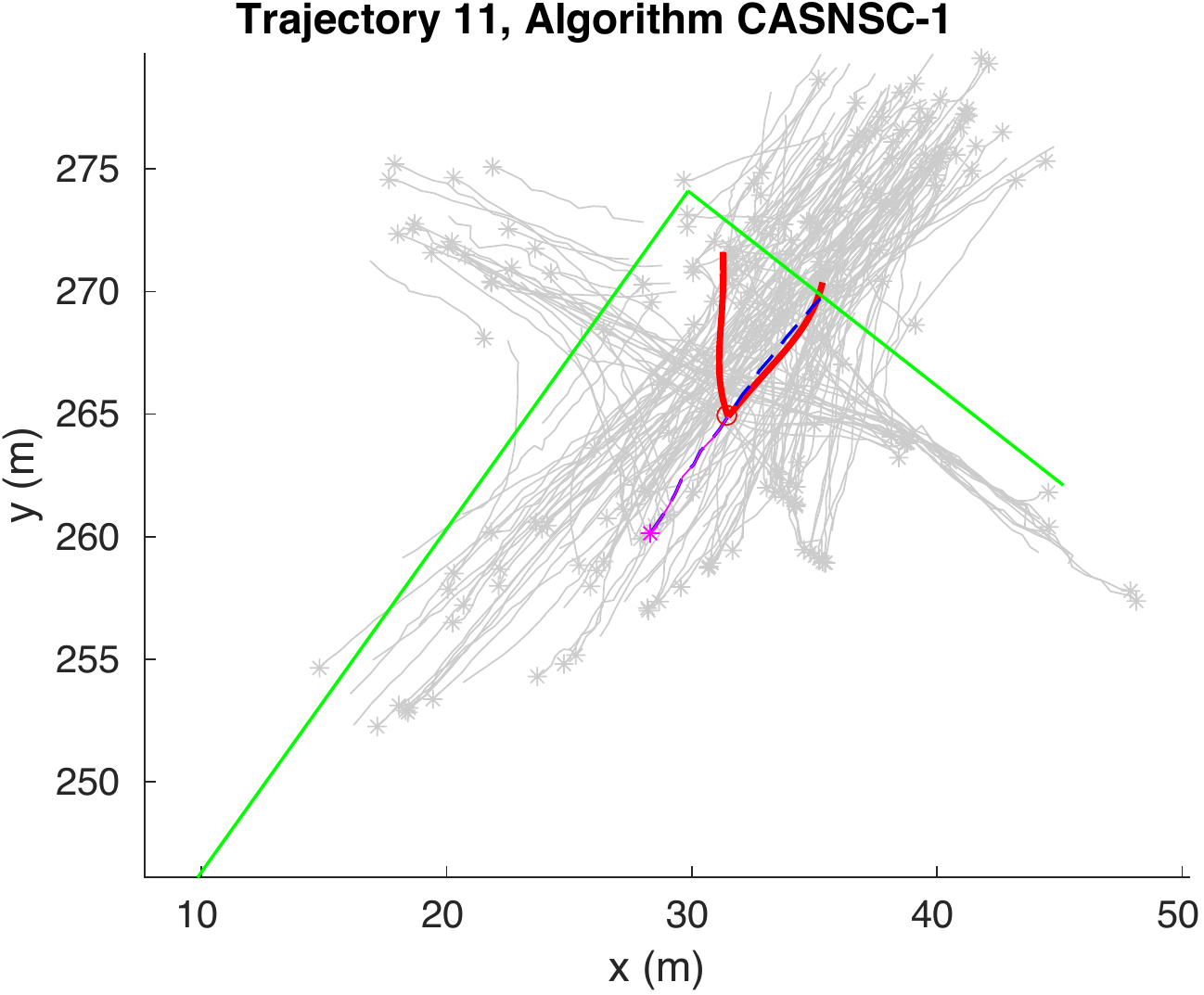}
\includegraphics[width=0.245\linewidth]{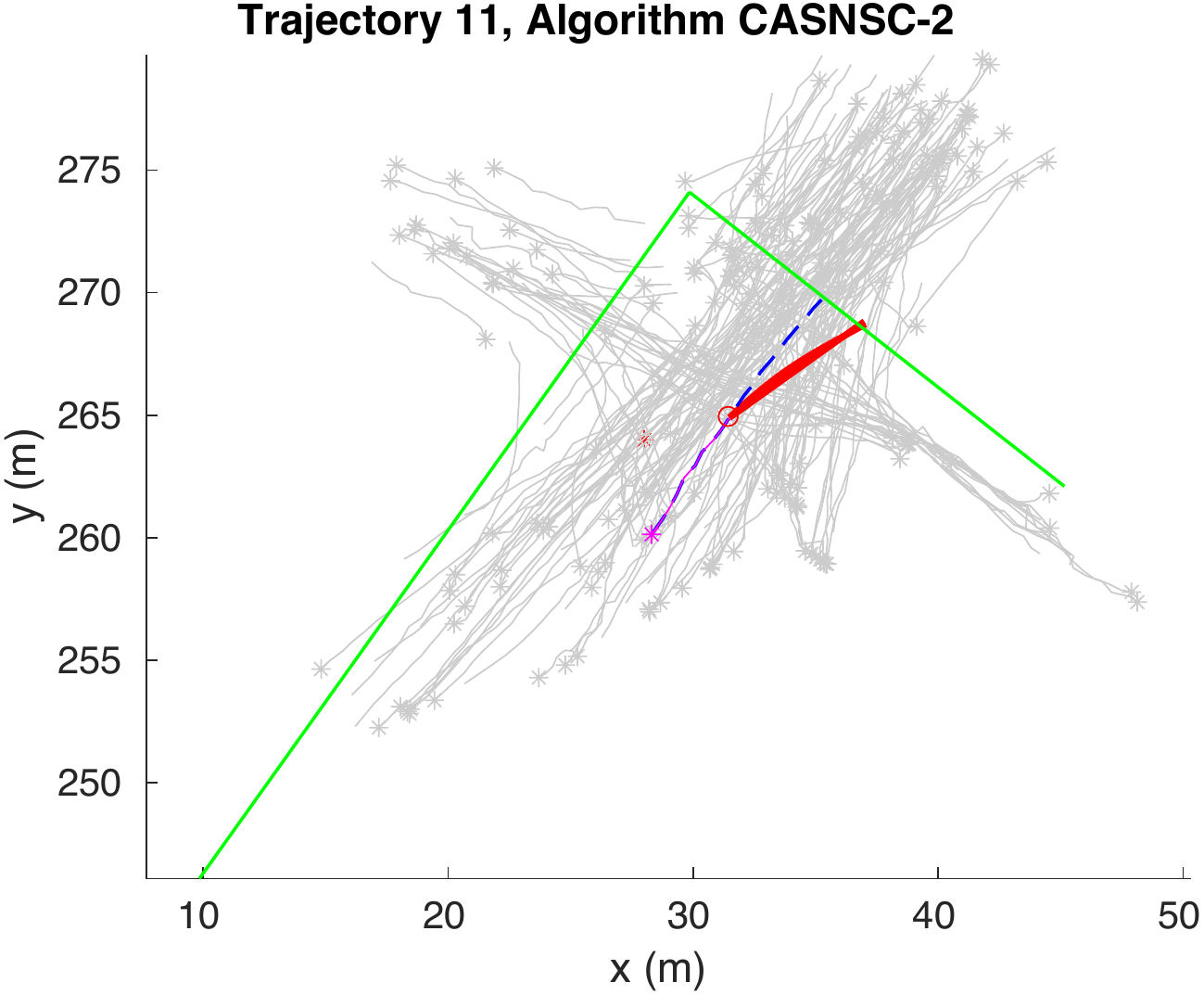}
\includegraphics[width=0.245\linewidth]{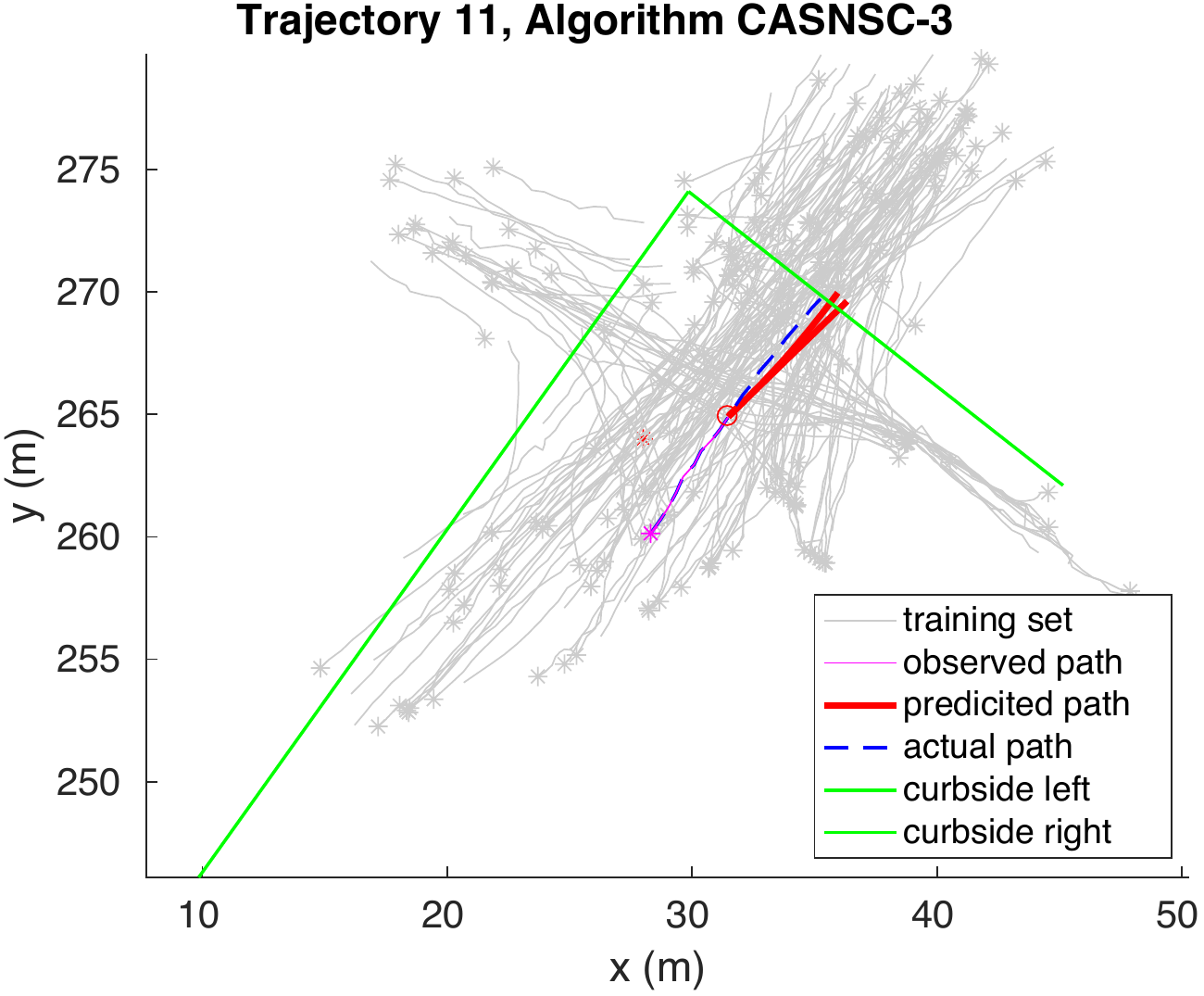}}
\subfigure[]{
\includegraphics[width=0.245\linewidth]{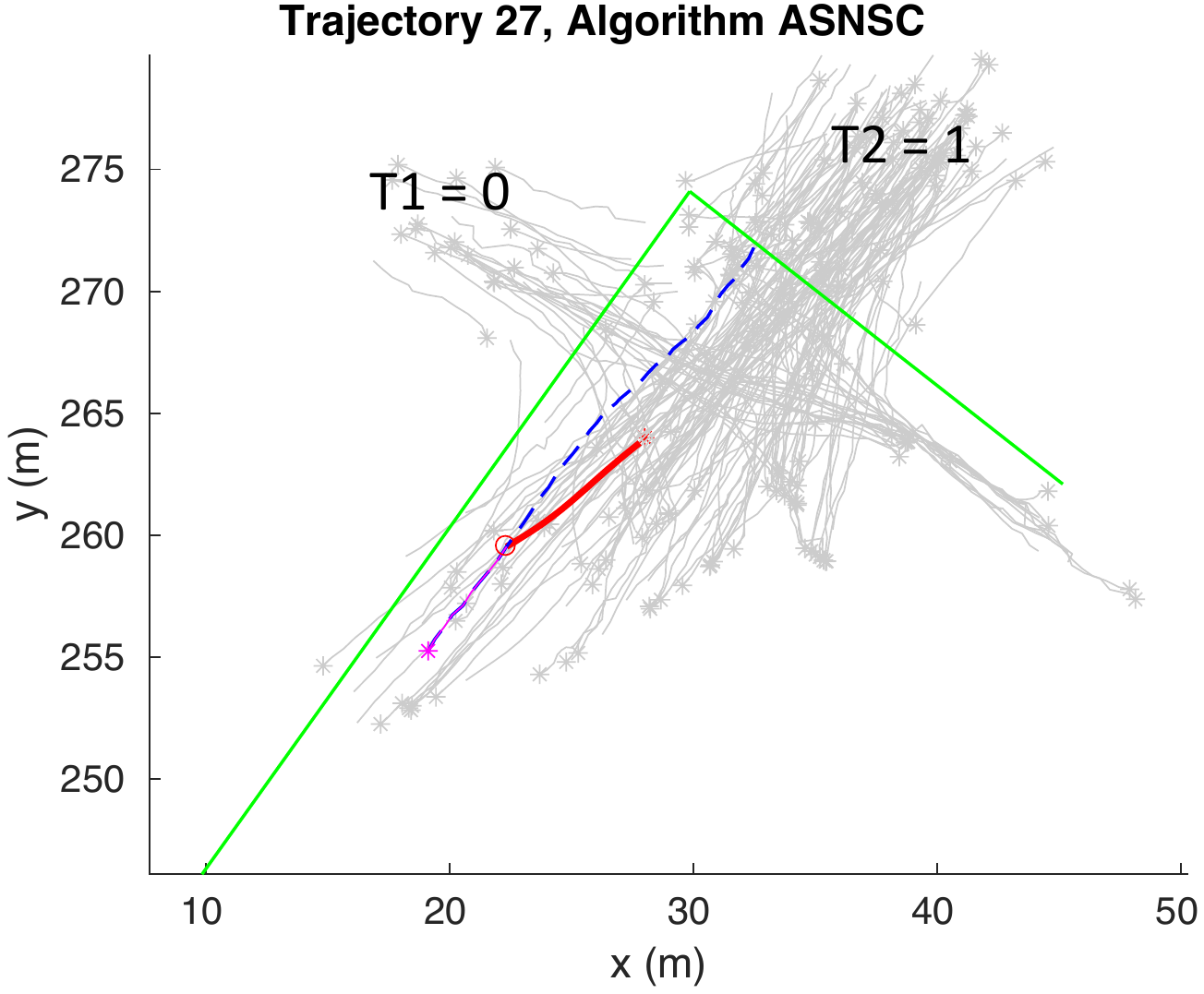}
\includegraphics[width=0.245\linewidth]{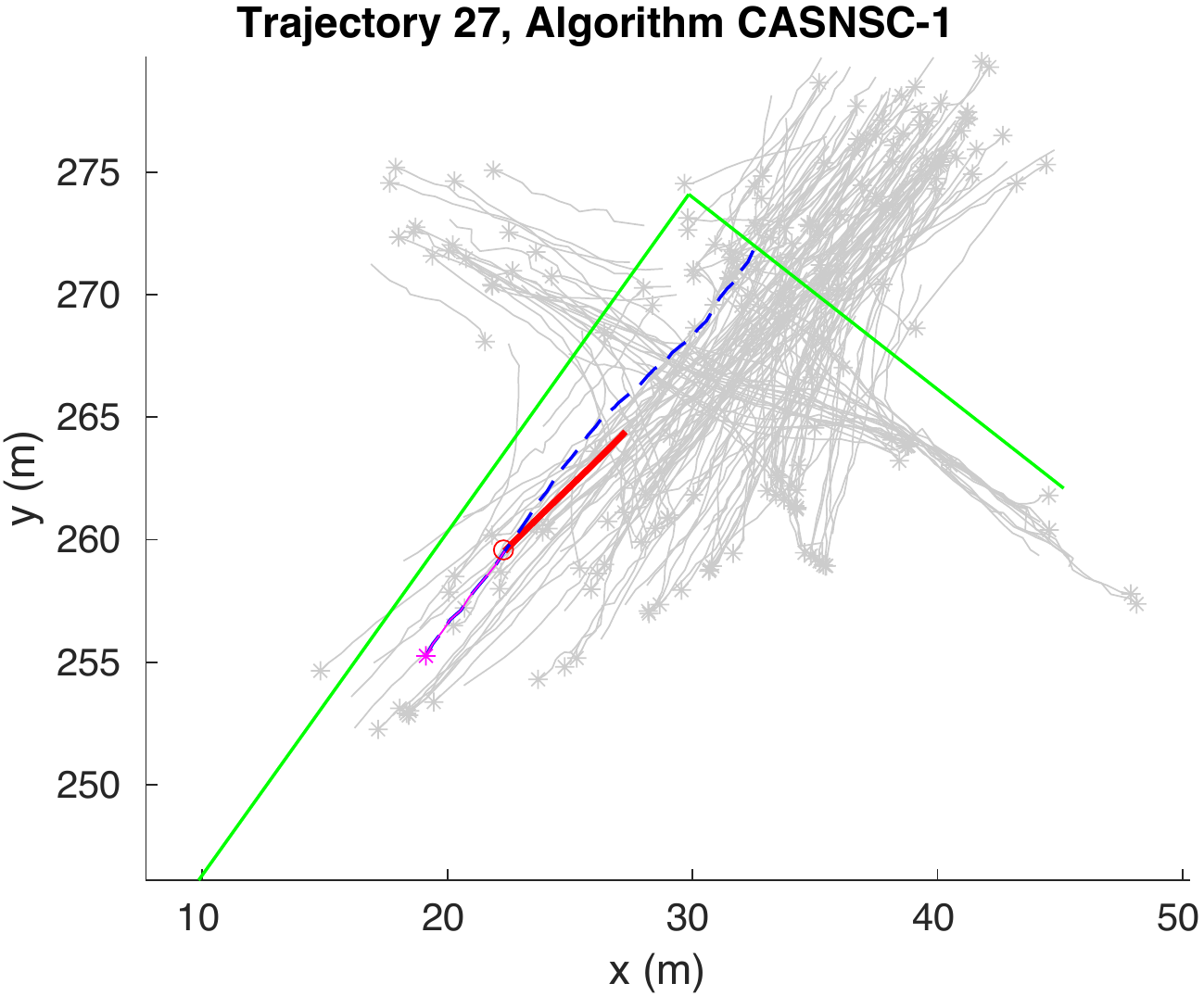}
\includegraphics[width=0.245\linewidth]{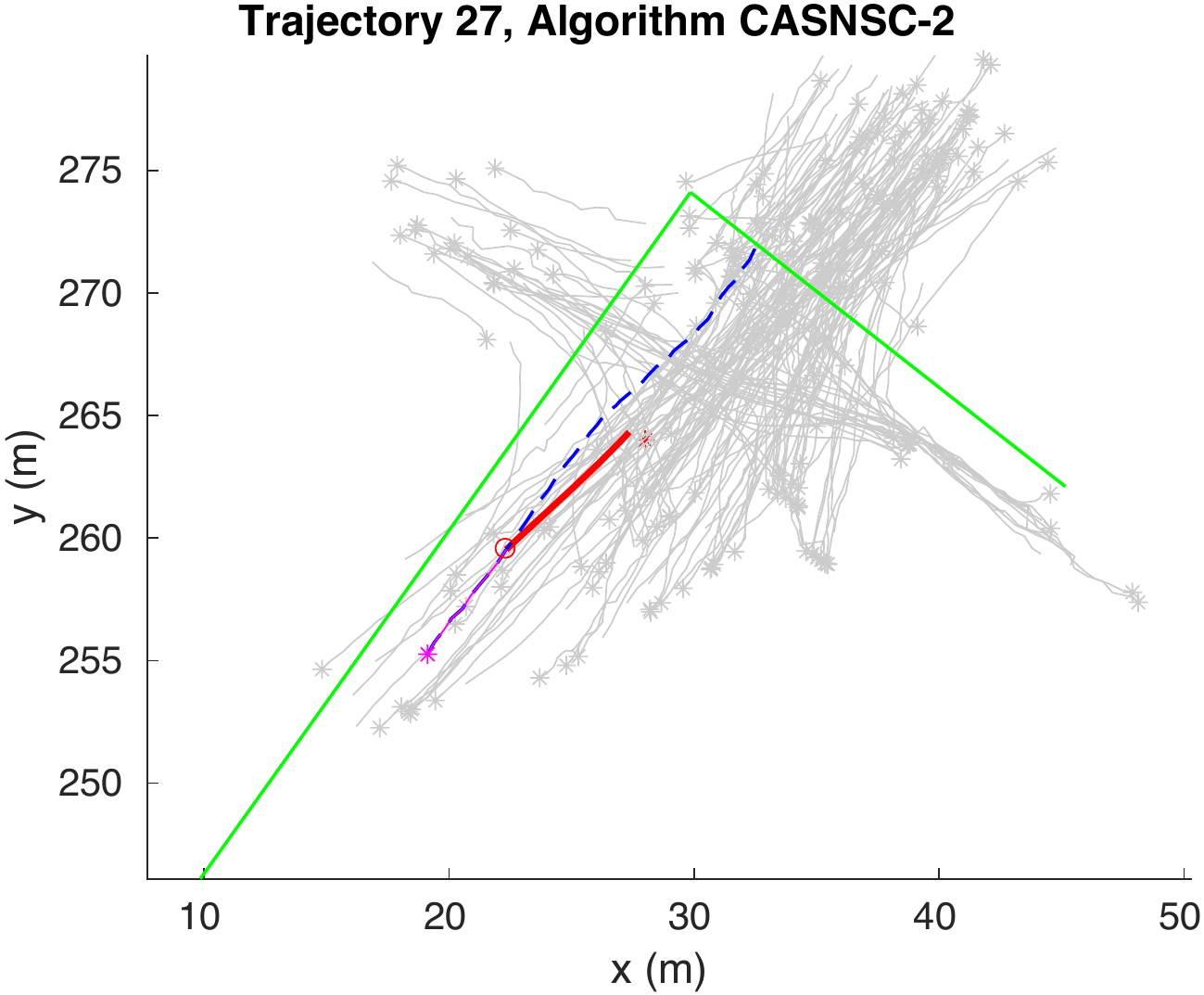}
\includegraphics[width=0.245\linewidth]{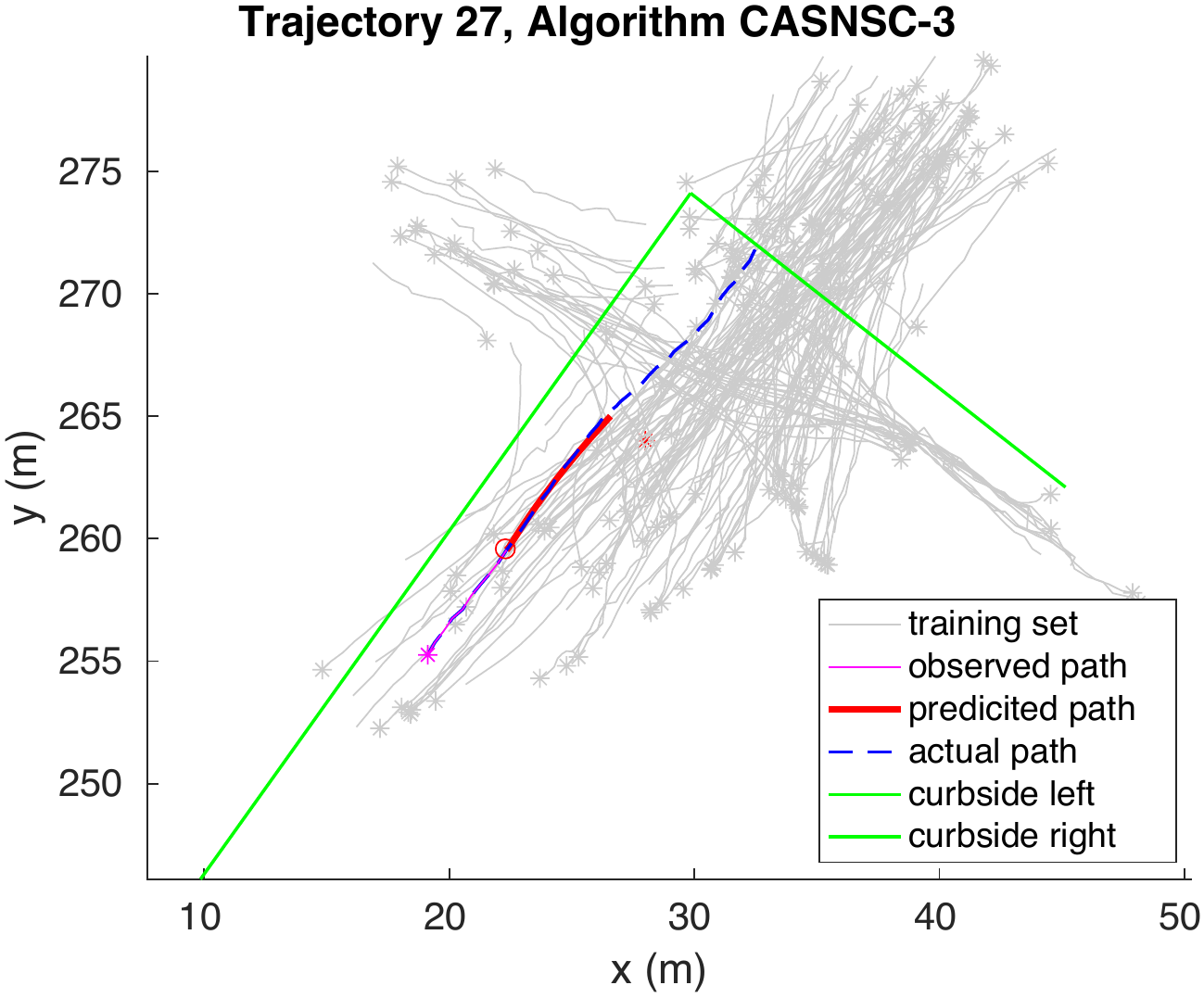}}
\subfigure[]{
\includegraphics[width=0.245\linewidth]{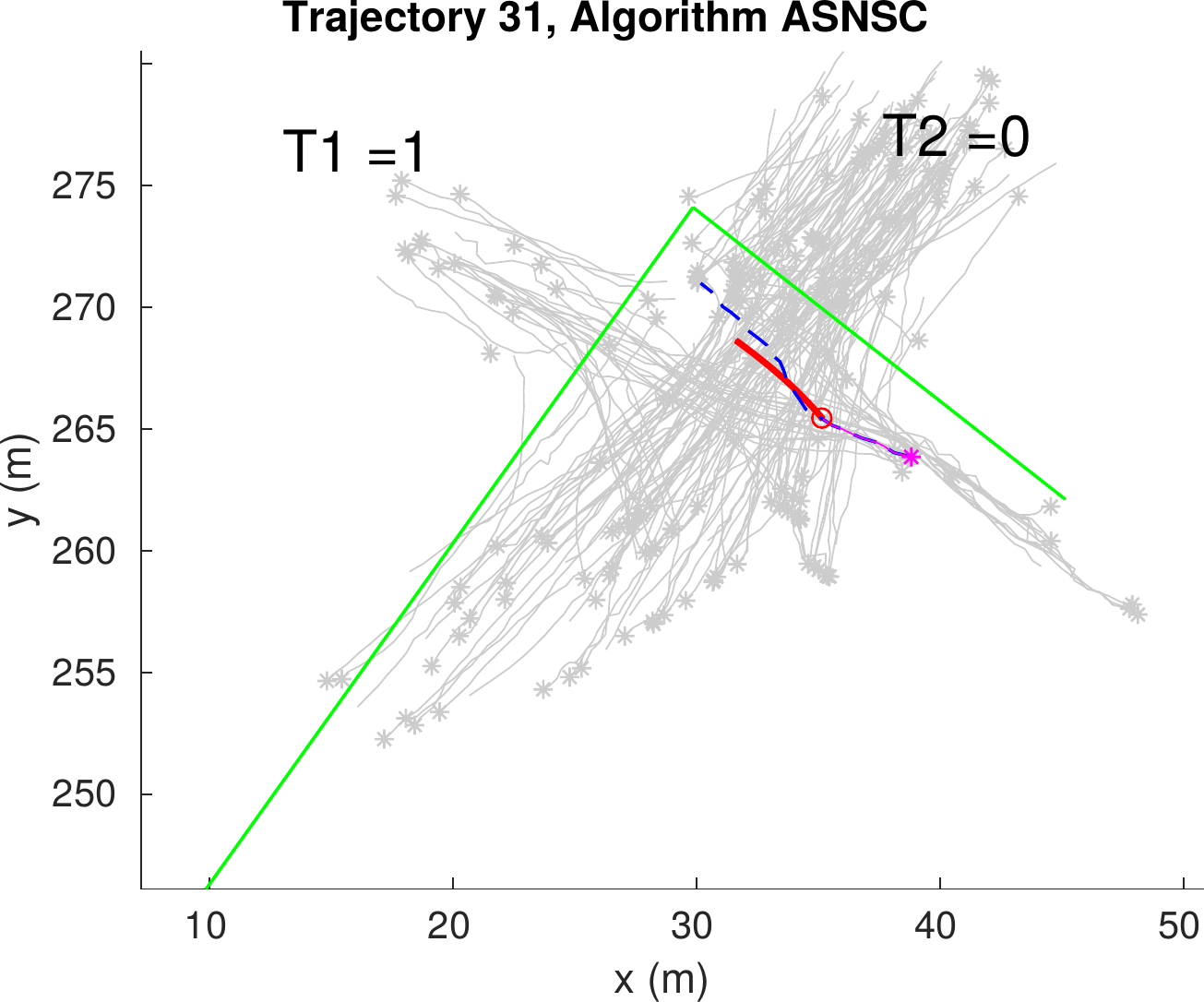}
\includegraphics[width=0.245\linewidth]{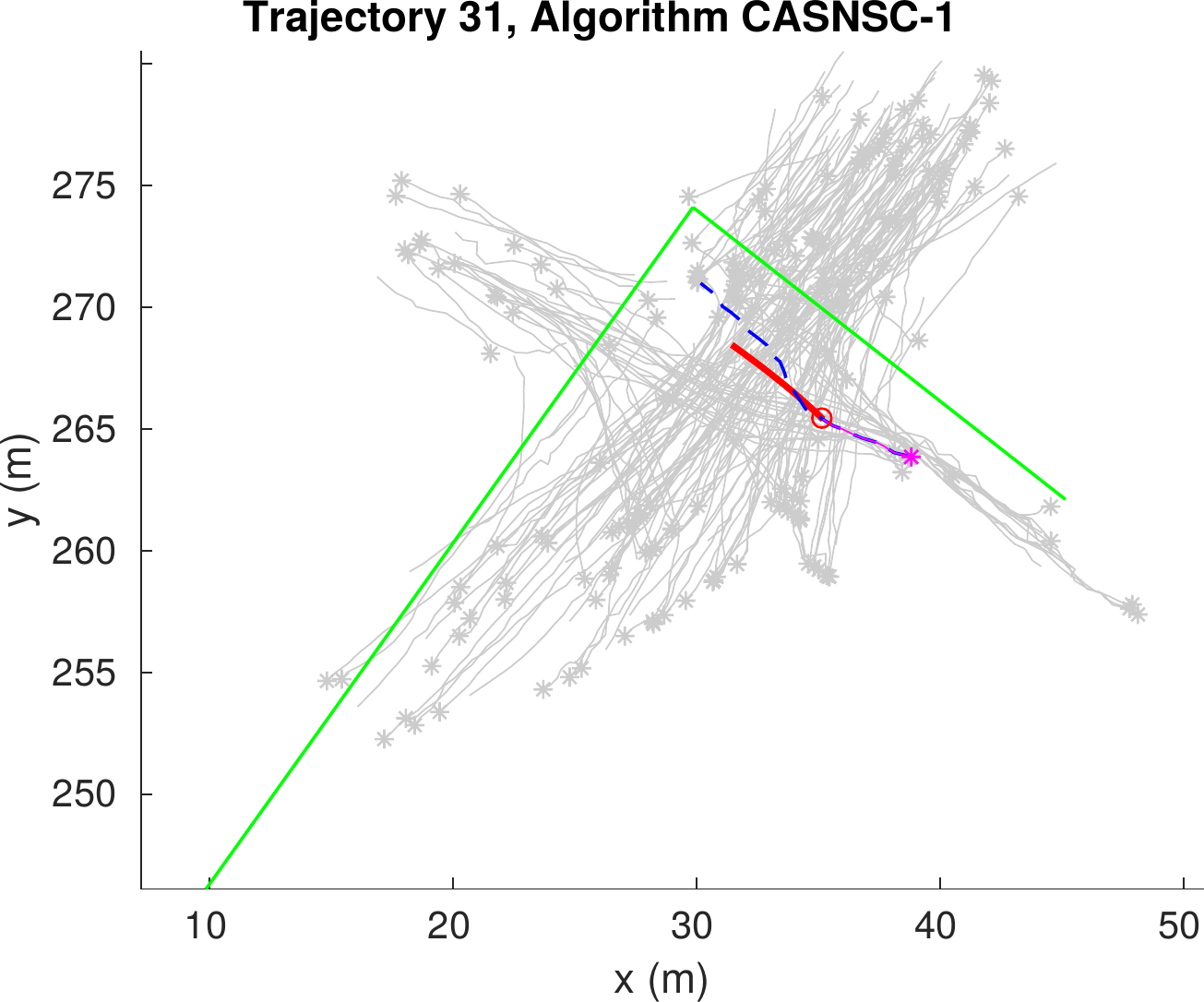}
\includegraphics[width=0.245\linewidth]{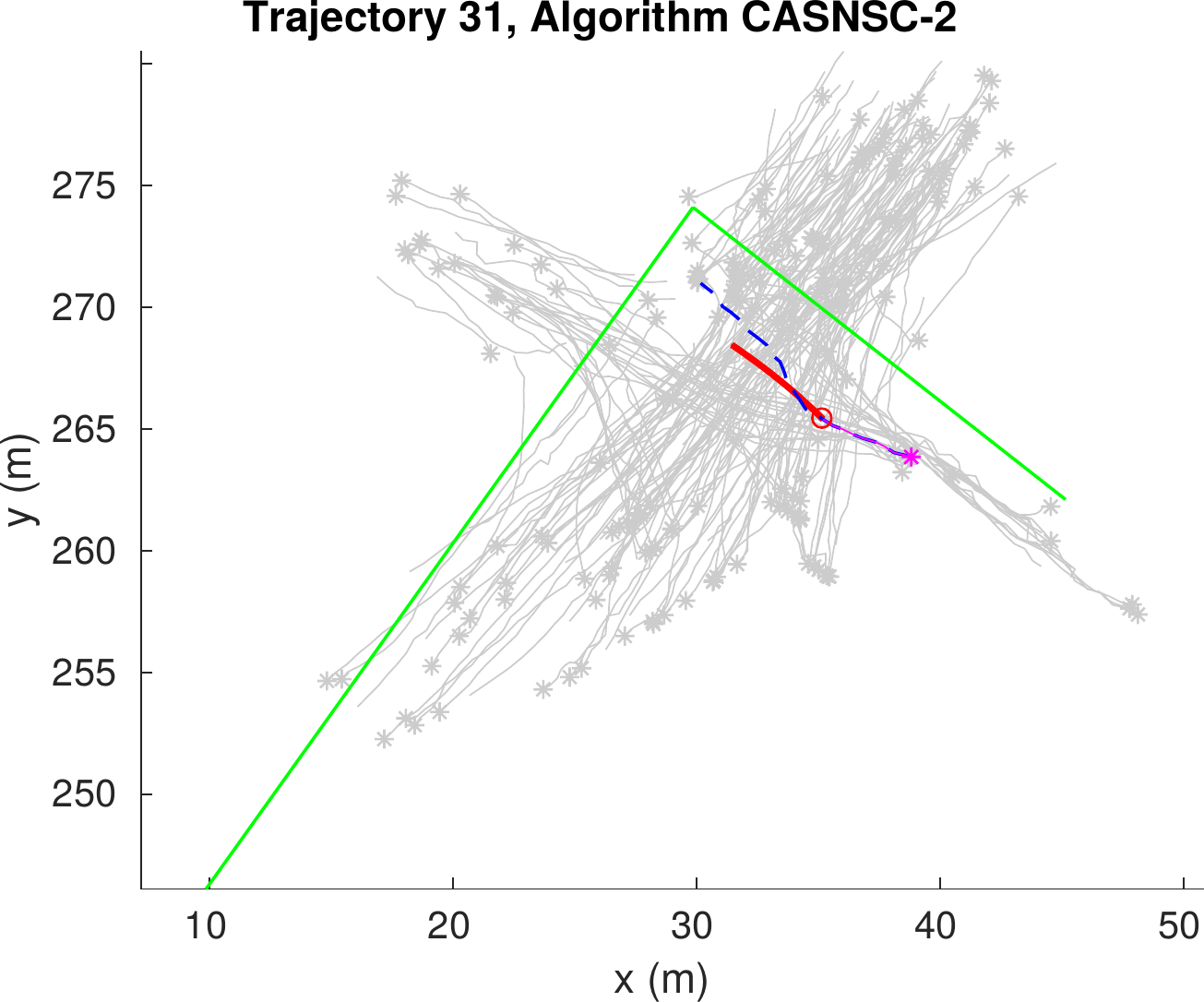}
\includegraphics[width=0.245\linewidth]{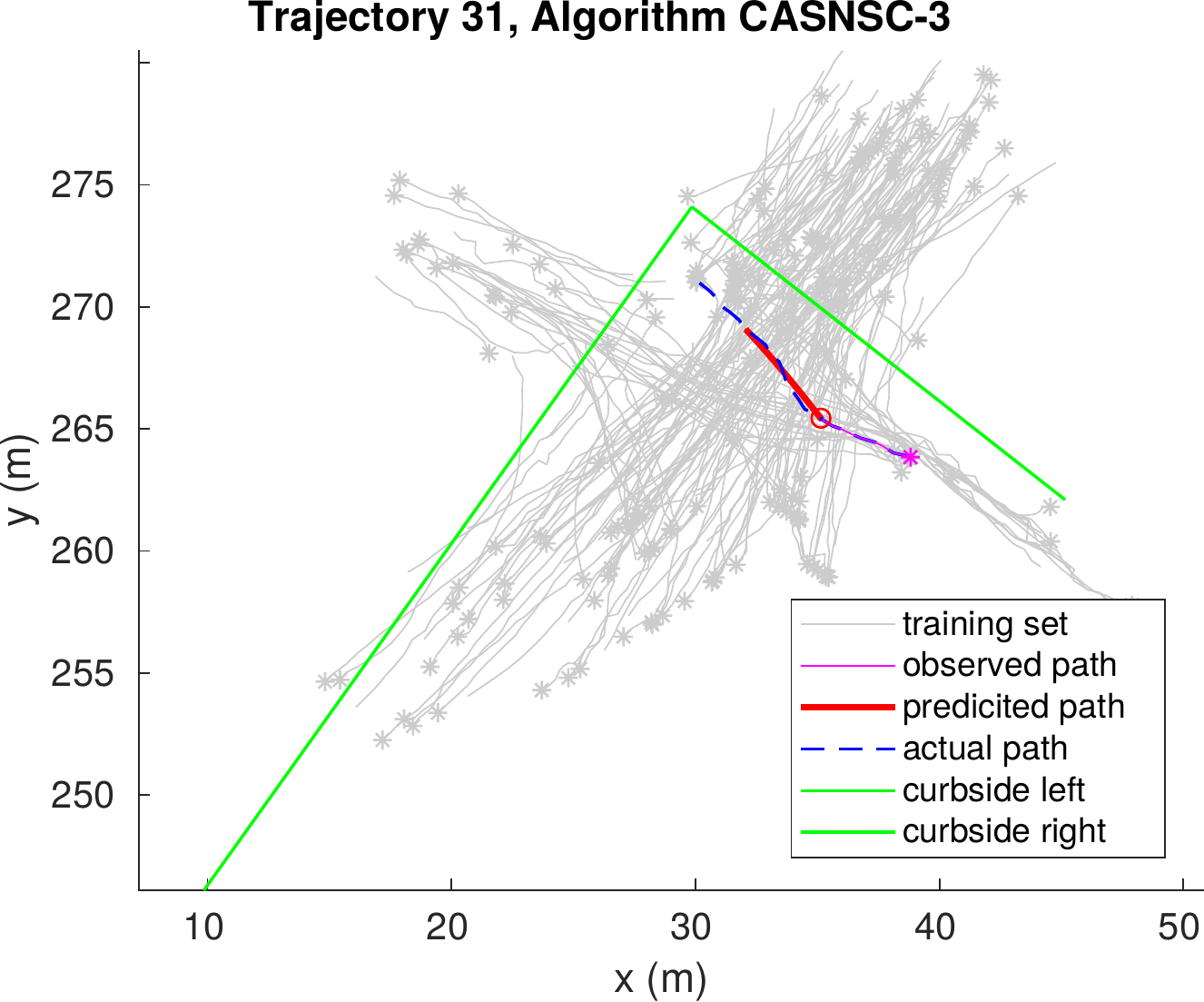}}
\caption{
  \label{fig:results_all}
  \small Comparison of prediction results of ASNSC (first column) with those of CASNSC-1: $\mathbf{X_t} = (x,y,tr)^T$, CASNSC-2: $\mathbf{X_t} = (x',y',tr)^T$ and CASNSC-3: $\mathbf{X_t} = (c_l,c_r,tr)^T$ (in the second, third and fourth columns respectively). Each row represents a different test trajectory. The curbside is shown using green, training trajectories using gray, observed path using pink, actual future path using dotted blue and predicted path using red lines.}
\end{figure*}

Our approach is tested on real pedestrian data collected by a Polaris GEM vehicle equipped with three Logitech C920 cameras and a SICK LMS151 LIDAR~\cite{miller2017predictive, miller2016dynamic}. The dataset consists of 218 training trajectories. A prior map of the environment is used to extract curbside boundaries. Pedestrian traffic light status is manually annotated. An observation history of 2.5 seconds prior to the pedestrian entering the intersection is used to predict 5 seconds ahead in time. Fig.~\ref{fig:results_all} provides a qualitative comparison of our approach with~\cite{chen2016augmented} using all 3 feature sets described in the previous section. While ASNSC provides all feasible pedestrian trajectories, given the intersection geometry, CASNSC picks those that are closest to the actual trajectory, in the correct direction, taking the context into account.

In the first scenario (trajectory 11), the \emph{pedestrian traffic lights'} status is given by T1 = 0, T2 = 1. The pedestrian enters the intersection and is faced with a choice between continuing to move straight or turn left. While ASNSC predicts a set of trajectories completely ignoring the context, CASNSC-1 is more confident of the future direction of motion as it can incorporate context (T2 = 1) into account. CASNSC-2 provides an even more confident prediction owing to the more accurate GP transition models created by incorporating \emph{curbside orientation}. CASNSC-3 outperforms all and its prediction is not just most confident but also follows the actual trajectory almost exactly. In the second scenario (trajectory 27), traffic light status is the same and while all four predictions are in the right direction, CASNSC-3 is again the most accurate. In the third scenario (trajectory 31), the traffic light status is given by T1 = 1, T2 = 0. Again, while all four predictions are in the right direction, CASNSC-3 is most accurate and follows the actual trajectory almost exactly. 

\begin{figure}[h]
  \begin{center}
    \includegraphics[trim=30 20 10 20,clip,width=.7\linewidth]{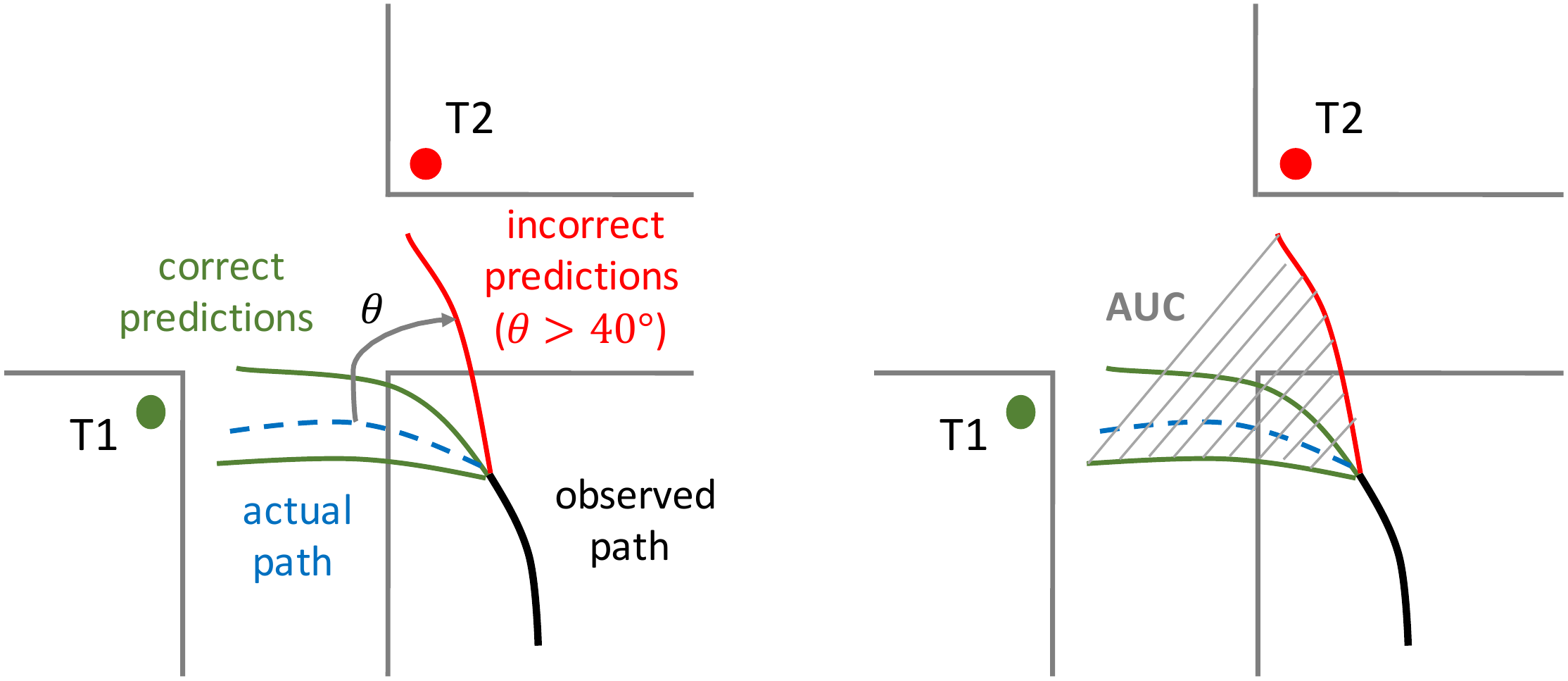}
  \end{center}
  \caption{\small(Left) \emph{Incorrect} and \emph{correct} predictions at an intersection scenario. (Right) Use of AUC as a metric for variance in prediction.}
  \label{fig:metric}
\end{figure}

Fig.~\ref{fig:metric} illustrates the metrics used for performance evaluation and Table~\ref{tab:comparison} provides a quantitative comparison of ASNSC with CASNSC-1, CASNSC-2 and CASNSC-3. As illustrated in Fig.~\ref{fig:metric}, the \emph{Area Under the Curve (AUC)} \cite{hand2009measuring} is used as a metric for comparing the confidence level of predictions, such that a larger \emph{AUC} corresponds to a lower confidence. Table~\ref{tab:comparison} indicates that \emph{AUC} for predictions using CASNSC-3 is the lowest, confirming that embedding context provides a more confident prediction. \emph{Classification accuracy} is also measured, which represents the fraction of \emph{correct} predictions, weighted by their likelihood for a more realistic estimate of the metric. Mathematically, if a set of $n$ trajectories is predicted as $\{\mathbf{t}_1,\hdots,\mathbf{t}_n\}$, with their likelihood of prediction given by $\{l_1,\hdots,l_n\}$, and the \emph{correct} predictions are identified as $\{\mathbf{t}_i\} \ \forall \ i \ \in \mathbf{C} \subset \{1,\hdots,n\}$, the \emph{classification accuracy} is given by:
\be
\text{Classification accuracy \%} = \frac{\sum_{i \in \mathbf{C}} l_i}{\sum_{k=1}^{n} l_k} \times 100 \% .
\ee
As seen in Fig.~\ref{fig:metric}, \emph{correct} predictions are defined as those in which the angular deviation from the observed trajectory i.e. $\theta$ is less than 40 degrees. In addition to the illustrated metrics, the \emph{Modified Hausdorff distance (MHD)} \cite{dubuisson1994modified} is used to compare predicted pedestrian trajectories with the ground truth. We again use the likelihood of predicted trajectories to compute the weighted average of MHD for a more accurate quantification of the metric.

\begin{table}[h]
  \caption{Performance evaluation comparison of CASNSC with ASNSC}
  \label{tab:comparison}
  \centering
  \begin{tabular}{lllll}
    \toprule
    Algorithm & Classification& MHD($m$) & AUC($m^2$) & Computation \\
    & ~accuracy($\%$) & & & \quad time(s) \\
    \midrule
    ASNSC & \qquad 83.71 & \ 2.09 & \ 131.13 & \qquad \ \textbf{0.03} \\
    CASNSC-1 & \qquad 85.25 & \ 2.33 & \ 105.50 & \qquad \ 0.51 \\  
    CASNSC-2 & \qquad 90.00 & \ 2.05 & \ 85.23 & \qquad \ 0.48 \\ 
    CASNSC-3 & \qquad \textbf{94.20} & \ \textbf{1.77} & \ \textbf{49.44} & \qquad \ 0.04 \\ 
    \bottomrule
  \end{tabular}
\end{table}

Table~\ref{tab:comparison} shows an improvement in all the chosen metrics, with only a slight increase in computation time. All computations were performed on an Intel Core i7-7700HQ processor in Matlab R2016b.
\section{CONCLUSION}

We extend ASNSC by incorporating semantic features from the environment to learn the transition between motion primitives of pedestrian trajectories (dictionary atoms) for more confident and accurate prediction. The results are presented using three different feature sets for embedding context into our model: pedestrian position \& pedestrian traffic light (CASNSC-1), curbside orientation \& pedestrian traffic light (CASNSC-2) and relative distance to curbside \& pedestrian traffic light (CASNSC-3). CASNSC-3 which uses a combination of relative distance to curbside and pedestrian traffic light as emph{transition features} shows a 12.5\% improvement in \emph{classification accuracy}, 15.3\% improvement in \emph{MHD} and reduces variance in prediction, as measured by \emph{AUC}, by a factor of 2.65. There is scope for further improvement on incorporation of other features like crosswalks, location of subway stations etc. Automatic scene understanding and feature learning as well as incorporating interaction between pedestrians will be parts of future work.

Using context features like relative distance to curbside provides geometric information which can be shown as invariant to a specific intersection scenario. This insight can be useful in knowledge transfer from one environment to the another. Making the prediction model flexible enough to be applied to similar but new environments unseen in the training phase will also be part of future work.

In this paper, we do not build on the sparse coding based dictionary learning part of~\cite{chen2016augmented} to embed context. However, more recently, papers published on \emph{deep dictionary learning}~\cite{tariyal2016greedy,stevens2017tensor} provide interesting insights and approaches to build on it as well for future work.

\section*{ACKNOWLEDGMENT}
The authors would like to thank Justin Miller, Michael Everett and Rohan Banerjee for all their help and support in data collection. This work is supported by a research grant from the Ford Motor Company.


\bibliographystyle{IEEEtran}
\bibliography{ref}

\begin{thebibliography}{10}
\providecommand{\url}[1]{#1}
\csname url@rmstyle\endcsname
\providecommand{\newblock}{\relax}
\providecommand{\bibinfo}[2]{#2}
\providecommand\BIBentrySTDinterwordspacing{\spaceskip=0pt\relax}
\providecommand\BIBentryALTinterwordstretchfactor{4}
\providecommand\BIBentryALTinterwordspacing{\spaceskip=\fontdimen2\font plus
\BIBentryALTinterwordstretchfactor\fontdimen3\font minus
  \fontdimen4\font\relax}
\providecommand\BIBforeignlanguage[2]{{%
\expandafter\ifx\csname l@#1\endcsname\relax
\typeout{** WARNING: IEEEtran.bst: No hyphenation pattern has been}%
\typeout{** loaded for the language `#1'. Using the pattern for}%
\typeout{** the default language instead.}%
\else
\language=\csname l@#1\endcsname
\fi
#2}}

\bibitem{chen2016augmented}
Y.~F. Chen, M.~Liu, and J.~P. How, ``Augmented dictionary learning for motion
  prediction,'' in \emph{Robotics and Automation (ICRA), 2016 IEEE
  International Conference on}.\hskip 1em plus 0.5em minus 0.4em\relax IEEE,
  2016, pp. 2527--2534.

\bibitem{miller2017predictive}
J.~Miller and J.~P. How, ``Predictive positioning and quality of service
  ridesharing for campus mobility on demand systems,'' in \emph{Robotics and
  Automation (ICRA), 2017 IEEE International Conference on}.\hskip 1em plus
  0.5em minus 0.4em\relax IEEE, 2017, pp. 1402--1408.

\bibitem{fagnant2015preparing}
D.~J. Fagnant and K.~Kockelman, ``Preparing a nation for autonomous vehicles:
  opportunities, barriers and policy recommendations,'' \emph{Transportation
  Research Part A: Policy and Practice}, vol.~77, pp. 167--181, 2015.

\bibitem{Bagloee2016}
\BIBentryALTinterwordspacing
S.~A. Bagloee, M.~Tavana, M.~Asadi, and T.~Oliver, ``Autonomous vehicles:
  challenges, opportunities, and future implications for transportation
  policies,'' \emph{Journal of Modern Transportation}, vol.~24, no.~4, pp.
  284--303, Dec 2016. [Online]. Available:
  \url{https://doi.org/10.1007/s40534-016-0117-3}
\BIBentrySTDinterwordspacing

\bibitem{lefevre2014survey}
S.~Lef{\`e}vre, D.~Vasquez, and C.~Laugier, ``A survey on motion prediction and
  risk assessment for intelligent vehicles,'' \emph{Robomech Journal}, vol.~1,
  no.~1, p.~1, 2014.

\bibitem{rasmussen2006gaussian}
C.~E. Rasmussen and C.~K. Williams, \emph{Gaussian processes for machine
  learning}.\hskip 1em plus 0.5em minus 0.4em\relax MIT press Cambridge, 2006,
  vol.~1.

\bibitem{kooij2014context}
J.~F.~P. Kooij, N.~Schneider, F.~Flohr, and D.~M. Gavrila, ``Context-based
  pedestrian path prediction,'' in \emph{European Conference on Computer
  Vision}.\hskip 1em plus 0.5em minus 0.4em\relax Springer, 2014, pp. 618--633.

\bibitem{bissacco2009hybrid}
A.~Bissacco and S.~Soatto, ``Hybrid dynamical models of human motion for the
  recognition of human gaits,'' \emph{International journal of computer
  vision}, vol.~85, no.~1, pp. 101--114, 2009.

\bibitem{gonzalez2004context}
A.~J. Gonzalez, W.~J. Gerber, R.~F. DeMara, and M.~Georgiopoulos,
  ``Context-driven near-term intention recognition,'' \emph{The Journal of
  Defense Modeling and Simulation}, vol.~1, no.~3, pp. 153--170, 2004.

\bibitem{goldhammer2013early}
M.~Goldhammer, M.~Gerhard, S.~Zernetsch, K.~Doll, and U.~Brunsmann, ``Early
  prediction of a pedestrian's trajectory at intersections,'' in
  \emph{Intelligent Transportation Systems-(ITSC), 2013 16th International IEEE
  Conference on}.\hskip 1em plus 0.5em minus 0.4em\relax IEEE, 2013, pp.
  237--242.

\bibitem{karasev2016intent}
V.~Karasev, A.~Ayvaci, B.~Heisele, and S.~Soatto, ``Intent-aware long-term
  prediction of pedestrian motion,'' in \emph{Robotics and Automation (ICRA),
  2016 IEEE International Conference on}.\hskip 1em plus 0.5em minus
  0.4em\relax IEEE, 2016, pp. 2543--2549.

\bibitem{alahi2016social}
A.~Alahi, K.~Goel, V.~Ramanathan, A.~Robicquet, L.~Fei-Fei, and S.~Savarese,
  ``Social lstm: Human trajectory prediction in crowded spaces,'' in
  \emph{Proceedings of the IEEE Conference on Computer Vision and Pattern
  Recognition}, 2016, pp. 961--971.

\bibitem{jacobs2017real}
H.~O. Jacobs, O.~K. Hughes, M.~Johnson-Roberson, and R.~Vasudevan, ``Real-time
  certified probabilistic pedestrian forecasting,'' \emph{IEEE Robotics and
  Automation Letters}, vol.~2, no.~4, pp. 2064--2071, 2017.

\bibitem{makris2002spatial}
D.~Makris and T.~Ellis, ``Spatial and probabilistic modelling of pedestrian
  behaviour,'' in \emph{British Machine Vision Conference 2002, vol. 2}.\hskip
  1em plus 0.5em minus 0.4em\relax Citeseer, 2002.

\bibitem{vasquez2009incremental}
D.~Vasquez, T.~Fraichard, and C.~Laugier, ``Incremental learning of statistical
  motion patterns with growing hidden markov models,'' \emph{IEEE Transactions
  on Intelligent Transportation Systems}, vol.~10, no.~3, pp. 403--416, 2009.

\bibitem{schulz2015controlled}
A.~T. Schulz and R.~Stiefelhagen, ``A controlled interactive multiple model
  filter for combined pedestrian intention recognition and path prediction,''
  in \emph{Intelligent Transportation Systems (ITSC), 2015 IEEE 18th
  International Conference on}.\hskip 1em plus 0.5em minus 0.4em\relax IEEE,
  2015, pp. 173--178.

\bibitem{rasmussen2002infinite}
C.~E. Rasmussen and Z.~Ghahramani, ``Infinite mixtures of gaussian process
  experts,'' in \emph{Advances in neural information processing systems}, 2002,
  pp. 881--888.

\bibitem{ferguson2015real}
S.~Ferguson, B.~Luders, R.~C. Grande, and J.~P. How, ``Real-time predictive
  modeling and robust avoidance of pedestrians with uncertain, changing
  intentions,'' in \emph{Algorithmic Foundations of Robotics XI}.\hskip 1em
  plus 0.5em minus 0.4em\relax Springer, 2015, pp. 161--177.

\bibitem{schulz2015pedestrian}
A.~T. Schulz and R.~Stiefelhagen, ``Pedestrian intention recognition using
  latent-dynamic conditional random fields,'' in \emph{Intelligent Vehicles
  Symposium (IV), 2015 IEEE}.\hskip 1em plus 0.5em minus 0.4em\relax IEEE,
  2015, pp. 622--627.

\bibitem{schneemann2016context}
F.~Schneemann and P.~Heinemann, ``Context-based detection of pedestrian
  crossing intention for autonomous driving in urban environments,'' in
  \emph{Intelligent Robots and Systems (IROS), 2016 IEEE/RSJ International
  Conference on}.\hskip 1em plus 0.5em minus 0.4em\relax IEEE, 2016, pp.
  2243--2248.

\bibitem{joseph2011bayesian}
J.~Joseph, F.~Doshi-Velez, A.~S. Huang, and N.~Roy, ``A bayesian nonparametric
  approach to modeling motion patterns,'' \emph{Autonomous Robots}, vol.~31,
  no.~4, p. 383, 2011.

\bibitem{aoude2013probabilistically}
G.~S. Aoude, B.~D. Luders, J.~M. Joseph, N.~Roy, and J.~P. How,
  ``Probabilistically safe motion planning to avoid dynamic obstacles with
  uncertain motion patterns,'' \emph{Autonomous Robots}, vol.~35, no.~1, pp.
  51--76, 2013.

\bibitem{miller2016dynamic}
J.~Miller, A.~Hasfura, S.-Y. Liu, and J.~P. How, ``Dynamic arrival rate
  estimation for campus mobility on demand network graphs,'' in
  \emph{Intelligent Robots and Systems (IROS), 2016 IEEE/RSJ International
  Conference on}.\hskip 1em plus 0.5em minus 0.4em\relax IEEE, 2016, pp.
  2285--2292.

\bibitem{hand2009measuring}
D.~J. Hand, ``Measuring classifier performance: a coherent alternative to the
  area under the roc curve,'' \emph{Machine learning}, vol.~77, no.~1, pp.
  103--123, 2009.

\bibitem{dubuisson1994modified}
M.-P. Dubuisson and A.~K. Jain, ``A modified hausdorff distance for object
  matching,'' in \emph{Pattern Recognition, 1994. Vol. 1-Conference A: Computer
  Vision \& Image Processing., Proceedings of the 12th IAPR International
  Conference on}, vol.~1.\hskip 1em plus 0.5em minus 0.4em\relax IEEE, 1994,
  pp. 566--568.

\bibitem{tariyal2016greedy}
S.~Tariyal, A.~Majumdar, R.~Singh, and M.~Vatsa, ``Greedy deep dictionary
  learning,'' \emph{arXiv preprint arXiv:1602.00203}, 2016.

\bibitem{stevens2017tensor}
A.~Stevens, Y.~Pu, Y.~Sun, G.~Spell, and L.~Carin, ``Tensor-dictionary learning
  with deep kruskal-factor analysis,'' in \emph{Artificial Intelligence and
  Statistics}, 2017, pp. 121--129.

\end{thebibliography}
\end{document}